\documentclass{segabs}
\usepackage{amsmath}
\usepackage{amsfonts} 
\usepackage{graphicx}
\usepackage{colortbl,booktabs}
\usepackage{float}
\usepackage{amsmath,amsthm,amssymb,amsfonts}
\usepackage{mathptmx}
\DeclareMathAlphabet{\mathcal}{OMS}{cmsy}{m}{n}
\DeclareSymbolFont{largesymbols}{OMX}{cmex}{m}{n}

\begin{document}

\title{3D seismic data denoising using two-dimensional sparse coding scheme}
\renewcommand{\thefootnote}{\fnsymbol{footnote}} 
\author{Ming-Jun Su$^{1}$, Jingbo Chang$^{2}$, Feng Qian$^{3}$, Guangmin Hu$^{3}$, Xiao-Yang Liu\footnotemark[1],\\
$^{1}$ PetroChina Research Institute of Petroleum Exploration and Development (RIPED)-Northwest,\\
$^{2}$ School of Communication and Information Engineering, University of Electronic Science and Technology of China,\\
$^{3}$ Center for Information Geoscience, University of Electronic Science and Technology of China,\\
\footnotemark Department of Electrical Engineering, Columbia University, NY, USA}

\footer{Example}
\lefthead{Dellinger \& Fomel}
\righthead{3D seismic data denoising using two-dimensional sparse coding scheme}

\maketitle
\begin{abstract}
Seismic data denoising is vital to geophysical applications and the transform-based function method is one of the most widely used techniques. However, it is challenging to design a suitable sparse representation to express a transform-based function group due to the complexity of seismic data. In this paper, we apply a seismic data denoising method based on learning-type overcomplete dictionaries which uses two-dimensional sparse coding (2DSC). First, we model the input seismic data and dictionaries as third-order tensors and introduce tensor-linear combinations for data approximation. Second, we apply learning-type overcomplete dictionary, i.e., optimal sparse data representation is achieved through learning and training. Third, we exploit the alternating minimization algorithm to solve the optimization problem of seismic denoising. Finally we evaluate its denoising performance on synthetic seismic data and land data survey. Experiment results show that the two-dimensional sparse coding scheme reduces computational costs and enhances the signal-to-noise ratio.
\end{abstract}
\section{Introduction}
Underground image construction and rock property estimation are important tasks for seismic exploration.
However, seismic data may be subject to too serious noise contamination to carry out further processing and interpretation. In many cases, therefore, noise mitigation is essential for extracting useful information from the raw measurements \citep{Zhai2014Seismic}.

Transform-based function method is one of the most commonly used techniques for seismic denoising.
This kind of methods commonly require the conversion of the data into different domains by using methods such as the Fourier transform, the Radon transform, and the curvelet transform \citep{Dashtian2015Denoising,Wang20133D}. 
However, the bases of the above transforms are pre-chosen and fixed during the process, so they lack of adaptability to dynamic data structures. \citet{Tang2012Seismic} introduce learning-type overcomplete dictionaries in which sparse data representation is achieved through learning and training the raw seismic data. As an effective dictionary learning algorithm, K-Singular Value Decomposition (K-SVD) has been applied to the problem of seismic denoising \citep{Beckouche2014Simultaneous}. However, the main drawback is its high computational complexity. In addition, as these algorithms mainly deal with vector-valued input data, the vectorization preprocess would break apart the local proximity and destruct the object structures of seismic data. \citet{Oropeza2011Simultaneous} present a rank reduction algorithm for noise attenuation of seismic records which adopts the multichannel singular spectrum analysis (MSSA). But this will cause strong residual noise and thus affect the following processing and interpretation tasks \citep{Chen2016An}. 

The basic idea that motivates us to address these challenges lies in two aspects: 1) tensor representation is able to preserve the local proximity and to capture the elementary object structures of seismic data; and 2) we exploit tensor-linear combinations to approximate the seismic data. Then the number of required bases can be significantly reduced, which can reduce the computational complexity.
\begin{figure}
\centering   
\includegraphics [height=3.9cm]{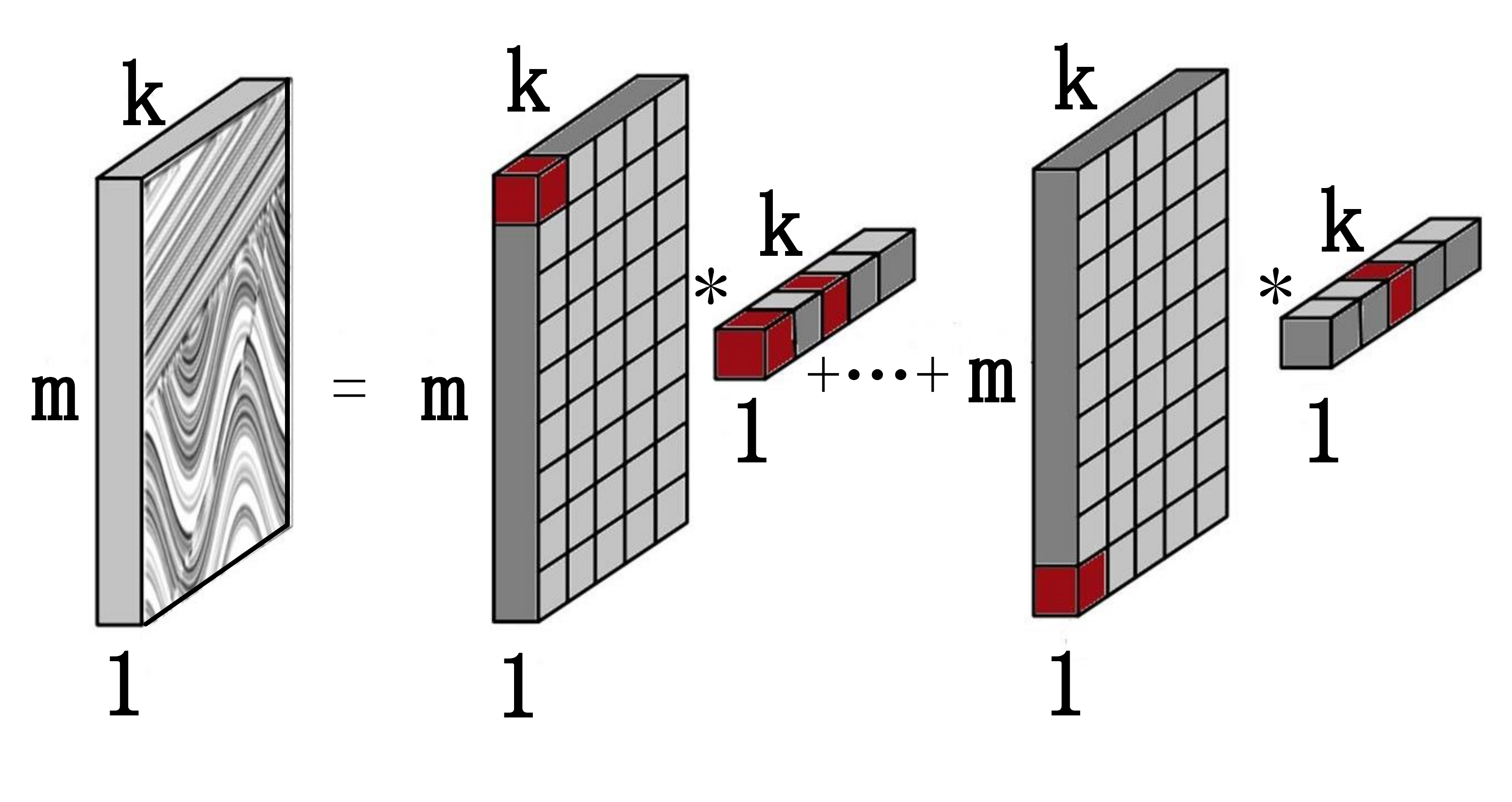}
\caption{Our model is based on the tensor-linear combination.} 
\label{fig:tensor-linearcombination}
\end{figure}

In this paper, we apply a seismic data denoising method based on learning-type overcomplete dictionaries which adopts the two dimensional sparse coding (2DSC) \citep{Jiang2017Graph,Liu2015Adaptive}. In the method, the input seismic data and dictionary are represented as third-order tensors, and the tensor-linear combinations  \citep{Kilmer2013Third} are used for data approximation. Figure 1 shows the tensor-linear combination. Alternating minimization algorithm which consists of a sparse coding learning step and a dictionary learning step can be used to solve the optimization problem. For sparse coding, we apply an iterative shrinkage thresholding algorithm \citep{Xu2014A} based on the tensor-product, which is directly implemented in the tensor space. For dictionary learning, we show that it can be solved efficiently by transforming to a Lagrange dual problem in the frequency domain.

Therefore, there are three major advantages of our scheme: 1) the size of dictionary can be significantly reduced  since the tensor-linear combination are used for seismic data approximation; 2) shifting invariant of tensor-product means that the seismic data can be generated by the shifted versions of bases without explicitly storing them; 3) the bases of dictionary can be trained and adaptively driven by target data, then it is able to develop dictionary transform-based functions suitable for special seismic signals.
\section*{NOTATION AND PRELIMINARY}
A third-order tensor is denoted as $\mathcal{Y} \in \mathbb{R}^{m\times n\times k}$. The discrete Fourier transform (DFT) along the third dimension of $\mathcal{Y}$ is denoted as $\widehat{\mathcal{Y}}$. The $\ell_1$-norm and Frobenius norm are denoted as $ \left \| \mathcal{Y}  \right \|_1=\sum _{i,\,j,\,k}\left | \mathcal{Y} (i,\,j,\,k) \right | $, and $\left \| \mathcal{Y}  \right \|_F=(\,\sum _{i,\,j,\,k} \sqrt{\mathcal{Y} (i,\,j,\,k)^2} \,)$. The superscript $H$ of $\mathcal{D}^{(l)^H}$ represents the conjugate transpose. Furthermore, we need two definitions.

\textbf{Definition1}: The tensor-product $\mathcal{Y}=\mathcal{D}*\mathcal{X}$ of $\mathcal{D}\in \mathbb{R}^{m\times r\times k}$ and $\mathcal{X}\in \mathbb{R}^{r\times n\times k}$ is a tensor  $\mathcal{Y}\in \mathbb{R}^{(m\times n\times k)}$, where $\mathcal{Y}(i\,,j,\,:)=\sum _{q=1}^{r}\mathcal{D}(i\,,q,\,:)*\mathcal{X}(q\,,j,\,:)$ and the $*$ denotes the circular convolution operation.

\textbf{Definition2}: The tensor-linear combination of the tensor bases $\left \{ \mathcal{D}_j \right \}_{j=1}^{r}\in \mathbb{R}^{m\times 1\times k}$ with the corresponding tensor coefficient $\left \{ \mathcal{X}_j \right \}_{j=1}^{r}\in \mathbb{R}^{1\times 1\times k}$ are defined as:
\begin{equation}
\mathcal{D}_1*\mathcal{X}_1+\cdots +\mathcal{D}_r*\mathcal{X}_r=\mathcal{D}*\mathcal{X},
\end{equation}
where $\mathcal{D}\in \mathbb{R}^{(m\times r\times k)}$ with $\mathcal{D}(:\,,j,\,:)=\mathcal{D}_j$, and $\mathcal{X}\in \mathbb{R}^{(m\times r\times k)}$.
\section*{Problem Formulation}
We use the following model to describe the noisy seismic data:
\begin{equation}
\mathcal{Y}=\mathcal{M}+\mathcal{\varepsilon}, 
\end{equation}
where $\mathcal{M}$ denotes the clean seismic data to be estimated, $\mathcal{Y}$ is the noisy seismic data and $\varepsilon$ is the random noise. To reduce noise $\varepsilon$ is to estimate $\mathcal{M}$ as accurate as possible from the acquired noisy data $\mathcal{Y}$.

Since seismic data sets are usually so complex with different geometrical features, we use the learning-based sparse dictionary to represent them. This dictionary is overcomplete and its elements are called atoms. In this overcomplete space domain, we adaptively search for an optimal composition of base atoms to sparsely represent the target signal through learning and training \citep{Zhu2015Seismic}.

Divide the seismic data $\mathcal{M}$ into several sub-blocks $\mathcal{M}_{ij}$ $(i,\,j \in\Omega )$ with fixed size. Assume each sub-block $\mathcal{M}_{ij}$ can be represented by a linear combination of several atoms extracted from the fixed dictionary $\mathcal{D}$ and the corresponding coefficients are denoted by $\mathcal{X}_{ij}$. Then we can formulate the problem with a general form, to minimize the following function with three penalties \citep{Tang2012Seismic}:
\begin{equation}
\begin{split}
f(\widehat{\mathcal{D}},\,\widehat{\mathcal{X}},\,\widehat{\mathcal{M}})=&\lambda \left \| \mathcal{M}-\mathcal{Y} \right \|_2^2+\sum _{i,\,j}\mu _{ij}\left \| \mathcal{X}_{ij} \right \|_0+\\
&\sum _{i,\,j}\left \| \mathcal{D}^H\mathcal{X}_{ij}-\mathcal{M}_{ij} \right \|_2^2, 
\end{split}
\end{equation}
where $\lambda$ and $\mu_{ij}$ denote penalty functions weights. If given $\widehat{\mathcal{M}}$ and $\widehat{\mathcal{D}}$ we can deduce $\widehat{\mathcal{X}}$ by minimizing equation (3). Once $\widehat{\mathcal{X}}$ is deduced, then $\widehat{\mathcal{M}}=\mathcal{D}^H\widehat{\mathcal{X}}$ can be obtained and dictionary $\widehat{\mathcal{D}}$ can be also updated by minimizing function (3).\\

\rule{8cm}{0.5mm}
\textbf{Algorithm 1} Algorithm for 2DSC\\
\rule{8cm}{0.5mm}
\textbf{Input:} seismic data: $\mathcal{Y} \in \mathbb{R}^{m\times n\times k}$, the number of atoms: $r$, sparsity regularizer: $\beta >0$, maximum iterative steps: num,\\
01:\quad\textbf{Initialization:} Randomly initialize $\mathcal{D}\in \mathbb{R}^{(m\times r\times k)}$ , $\mathcal{X} := 0\in \mathbb{R}^{(m\times r\times k)}$, and Lagrange dual variables $\lambda$, set $\mathcal{C}_1=\mathcal{X}_0\in \mathbb{R}^{(r\times 1\times k)}$, $t_1=1$,\\
02:\quad\textbf{for} iter = 1 to num \textbf{do}\\
03:\qquad//\textbf{Tensor Coefficients Learning }\\
04:\qquad Set $L^P=\eta ^p(\sum_{l=1}^{k}\left \|\widehat{\mathcal{D}}^{(l)^H}\widehat{\mathcal{D}}^{(l)}  \right \|_F)$,\\
05:\qquad Compute $\bigtriangledown f(\mathcal{C}_p)$,\\
06:\qquad Compute $\mathcal{X}_p$ via $\mathbf{Prox}_{\beta /L_p}(\mathcal{C}_p-\frac{1}{L_p}\bigtriangledown f(\mathcal{C}_p))$, \\
07:\qquad $t_{p+1}=\frac{1+\sqrt{1+4t_p^2}}{2}$,\\
08:\qquad $\mathcal{C}_{p+1}=\mathcal{X}_p+\frac{t_{p-1}}{t_{p+1}}(\mathcal{X}_p-\mathcal{X}_{p-1})$,\\
09:\qquad//\textbf{Tensor Dictionary Learning}\\
10:\qquad$\widehat{\mathcal{Y}}=\text{fft}(\mathcal{Y},[\,],3)$, $\widehat{\mathcal{X}}=\text{fft}(\mathcal{X},[\,],3),$\\
11:\qquad \textbf{for} $l$ = 1 to $k$ do\\
12:\qquad\qquad Solving $\Lambda $ by Newton's method,\\
13:\qquad\qquad  $\widehat{\mathcal{D}}^{(l)}=(\widehat{\mathcal{Y}}^{(l)}\widehat{\mathcal{X}}^{(l)H})(\widehat{\mathcal{X}}^{(l)}\widehat{\mathcal{X}}^{(l)H}+\Lambda )^{-1}$,\\
14:\qquad\textbf{end for}\\
15:\qquad$\mathcal{D}=\text{ifft}(\widehat{\mathcal{D}}, [\,], 3)$,\\
16:\quad \textbf{end for}\\
\textbf{Output:} $\mathcal{D},\mathcal{X}.$\\
\rule{8cm}{0.5mm}
\section*{2DSC model}
In this section, we apply the two-dimensional sparse coding to dictionary learning algorithm. We represent the training sample of 3-D seismic data as a third-order tensor $\mathcal{Y} \in \mathbb{R}^{m\times n\times k}$, then the adaptive dictionary learning process can be described as the following problems:
\begin{equation}
\begin{split}
\underset{\mathcal{D},\,\mathcal{X}}{\text{min}}\qquad &\frac{1}{2}\left \| \mathcal{Y} -\mathcal{D}*\mathcal{X} \right \|_{F}^{2}+\beta \left \| \mathcal{X} \right \|_1,\\
\text{s.t. }\qquad &\left \| \mathcal{D}(:,j:)) \right \|_{F}^{2}\leq 1,\,j\in \left \{ 1,\,2,\,\cdots ,\,k \right \},
\end{split}
\end{equation}
where $\mathcal{D}\in \mathbb{R}^{(m\times r\times k)}$ is the tensor dictionary where each lateral slice $\mathcal{D}(:,j,:)=\mathcal{D}_j$ is a basis, $\mathcal{X}\in \mathbb{R}^{(r\times n\times k)}$ is the tensor coefficient. The parameter $\beta $ balances the approximation error and the sparsity of the tensor coefficients, and $r$ is the number of atoms. 

\subsection{Solution}

Problem (4) can be solved by an iterative method that alternates between sparse tensor coefficients of the examples based on the current dictionary and a process of updating the dictionary atoms to better fit the data (Algorithm 1). The update of the dictionary is combined with an update of the sparse representations, thereby
accelerating convergence. The sparse coding learning step and the dictionary learning step are showd as follows:
\begin{figure*}
\centering   
\includegraphics [height=5.8cm]{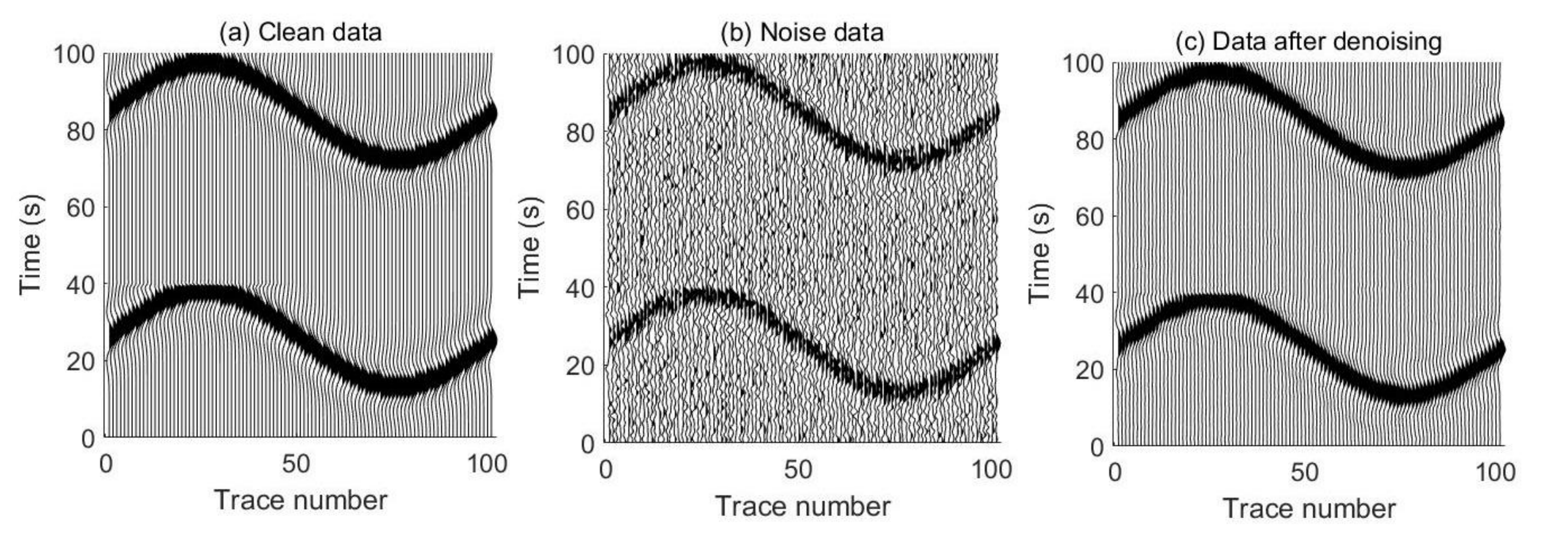}
\caption{Denoising analysis on synthetic data example.} 
\label{fig:Denoisinganalysis}
\end{figure*}

\subsubsection{Learning Tensor Coefficients}

Given the dictionary $\mathcal{D}\in \mathbb{R}^{(m\times r\times k)}$, solving the tensor sparse representations of seismic data $\mathcal{X}\in \mathbb{R}^{(r\times n\times k)}$ are converted to the following problem as:
\begin{equation}
\underset{\mathcal{X}\in \mathbb{R}^{(r\times n\times k)}}{\text{min}}\qquad \frac{1}{2}\left \| \mathcal{Y} -\mathcal{D}*\mathcal{X} \right \|_{F}^{2}+\beta \left \| \mathcal{X} \right \|_1.\\
\end{equation}
Rewrite (5) as:
\begin{equation}
\underset{\mathcal{X}\in \mathbb{R}^{(r\times n\times k)}}{\text{min}}\qquad f(\mathcal{X})+\beta g(\mathcal{X}),
\end{equation}
then (6) can be sloved by Iterative Shrinkage Thresholding algorithm based on Tensor-product (ISTA-T), which is similar to FISTA \citep{Beck2009A}. Then at the $p+1$-th iteration, $B_p+1$ can be updated by:
\begin{equation}
\begin{aligned}
\mathcal{X}_{p+1}=\text{arg}\,\underset{\mathcal{X}}{\text{min}}&f(\mathcal{X}_p)+<\bigtriangledown f(\mathcal{X}_p),\,\mathcal{X}-\mathcal{X}_p>+\\
&\frac{L_{p+1}}{2}\left \| \mathcal{X}-\mathcal{X}_p \right \|_F^2+\beta g(\mathcal{X}).
\end{aligned}
\end{equation}
\subsubsection{Tensor Dictionary Learning}

For learning the dictionary $\mathcal{D}$ while fixed $\mathcal{X}$, the optimization problem is:
\begin{equation}
\begin{split}
\underset{\mathcal{D}\in \mathbb{R}^{(m\times r\times k)}}{\text{min}}\qquad &\frac{1}{2}\left \| \mathcal{Y} -\mathcal{D}*\mathcal{X} \right \|_{F}^{2},\\
\text{s.t. }\qquad\quad &\left \| \mathcal{D}(:,j:)) \right \|_{F}^{2}\leq 1,\,j\in\left \{ 1,\,2,\,\cdots ,\,k \right \}.
\end{split}
\end{equation}
Decompose (8) into $k$ nearly-independent problems by DFT as follows:
\begin{equation}
\begin{split}
\underset{\widehat{\mathcal{D}}^{(l)}}{\text{min}}\qquad &\sum_{l=1}^{k}\left \| \widehat{\mathcal{Y} }^{(l)}-\widehat{\mathcal{D}}^{(l)}*\widehat{\mathcal{X}}^{(l)} \right \|_{F}^{2},\,l\in \left \{ 1,\,2,\,\cdots ,\,k \right \},\\
\text{s.t.}\qquad &\sum_{l=1}^{k}\left \| \widehat{\mathcal{D}}^{(l)}(:,j) \right \|_{F}^{2}\leq k,\,j\in\left \{ 1,\,2,\,\cdots ,\,k \right \}.
\end{split}
\end{equation}
Then, we adopt the Lagrange dual for solving (9) in frequency domain and the Lagrangian of (9) is:
\begin{equation}
\begin{split}
L(\widehat{\mathcal{D}},\Lambda )=&\sum_{l=1}^{r}\left \| \widehat{\mathcal{Y} }^{(l)}-\widehat{\mathcal{D}}^{(l)}*\widehat{\mathcal{X}}^{(l)} \right \|_{F}^{2}+\\
&\sum_{j=1}^{r}\lambda _j(\sum_{l=1}^{k}\left \| \widehat{\mathcal{D}}^{(l)}(:,j) \right \|_{F}^{2}-k),
\end{split}
\end{equation}
where $\lambda _j\geq 0$, $j\in \left \{ 1,\,2,\,\cdots ,\,k \right \}$ is a dual variable, and $\Lambda =diag(\lambda )$. The advantage of Lagrange dual is that the number of optimization variables is much smaller than which in the primal problem (8).
\section*{EVALUATION}
In this section, we use a synthetic example and a field data example to compare the denoising performance of the 2DSC scheme whih other two methods. To numerically test the performance and to quantify the comparison, we simulate noisy data by adding Gaussian white noise and define the criterion for comparison as signal-to-noise ratio (SNR):
\begin{equation}
\begin{split}
SNR=10\,\text{log}_{10}\frac{\left \| \mathcal{M} \right \|_2^2}{\left \|  \mathcal{M}-\mathcal{Y} \right \|_2^2}.
\end{split}
\end{equation}
Here, $\mathcal{M}$ is noise-free data, $ \mathcal{Y}$ is the denoised data.

\subsection{Synthetic data example}

To evaluate the method effectiveness, we use a Ricker wavelet with central frequency of 60 Hz to generate a simple 3-D model in which there are two dipping planes located around 300 and 900 m on the vertical axis, respectively. The clean and noisy data are shown in Figure 2(a) and 2(b), respectively. After applying the 2DSC method, the noise suppression section is shown in Figure 2(c). These three figures show that this method can effectively suppress noise. For comparison, we also used the K-SVD and MSSA methods to process the model. The comparison of SNR is shown in the first row of Table 1. The 2DSC can get better results considering the preservation of useful signals and removal of noise. And its calculated SNR also is the highest among all the three approaches.
\begin{table}[H]
\centering
\caption{Comparison of SNR using different approaches.} 
\begin{tabular}{ccccc}    
      \toprule[1pt]
      Models & Original & 2DSC & K-SVD & MSSA  \\
      \midrule
      Synthetic & 0.1403&13.8268 &10.2818& 9.5052\\
      Field &13.6809&26.2310 &24.0547 & 21.7438\\
      \bottomrule[1pt]
\end{tabular}
\end{table}
\begin{figure}[H]
\centering
\begin{minipage}{0.5\linewidth}
  \centerline{\includegraphics[width=7.3cm]{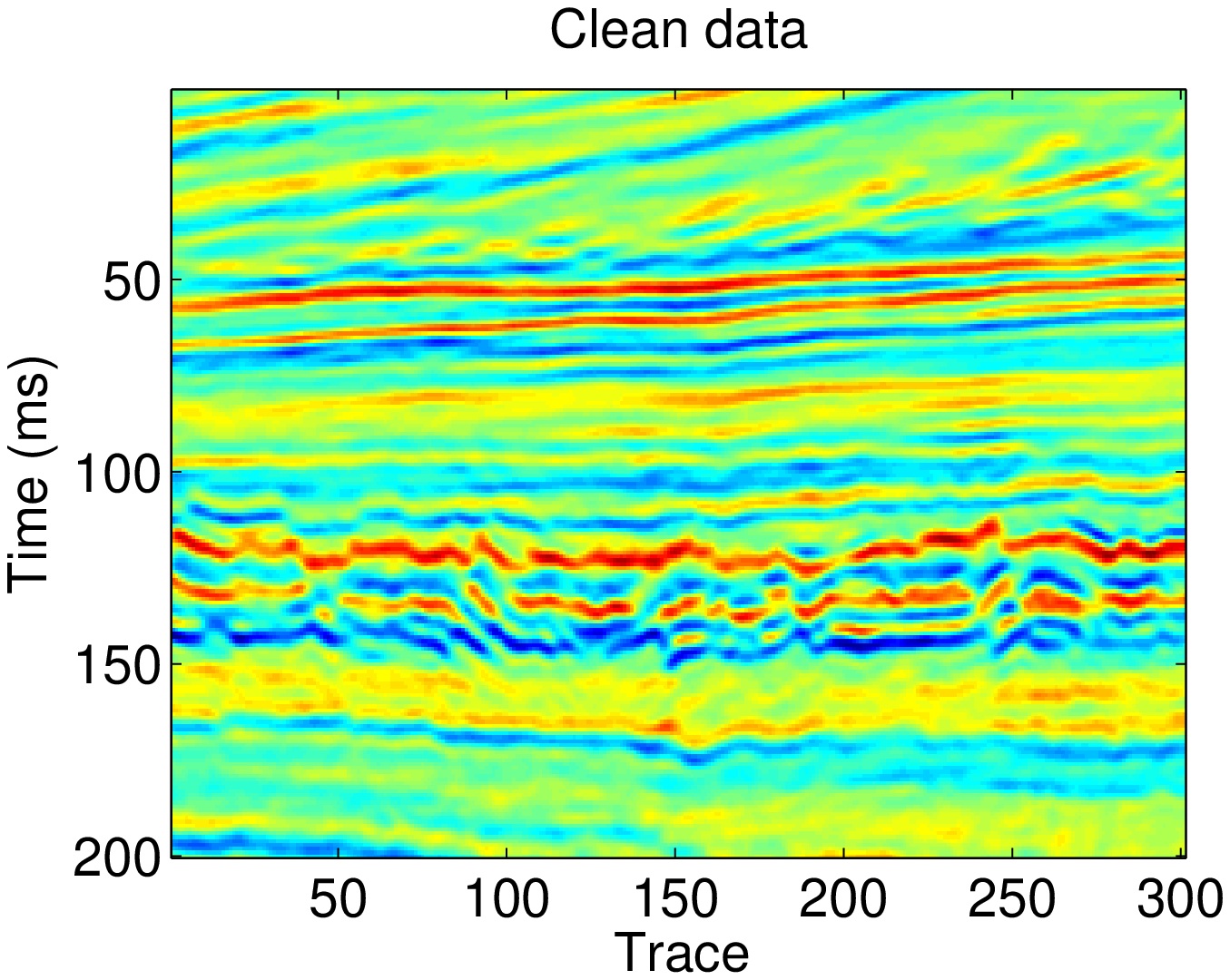}}
  \centerline{}
\end{minipage}
\begin{minipage}{0.5\linewidth}
  \centerline{\includegraphics[width=7.3cm]{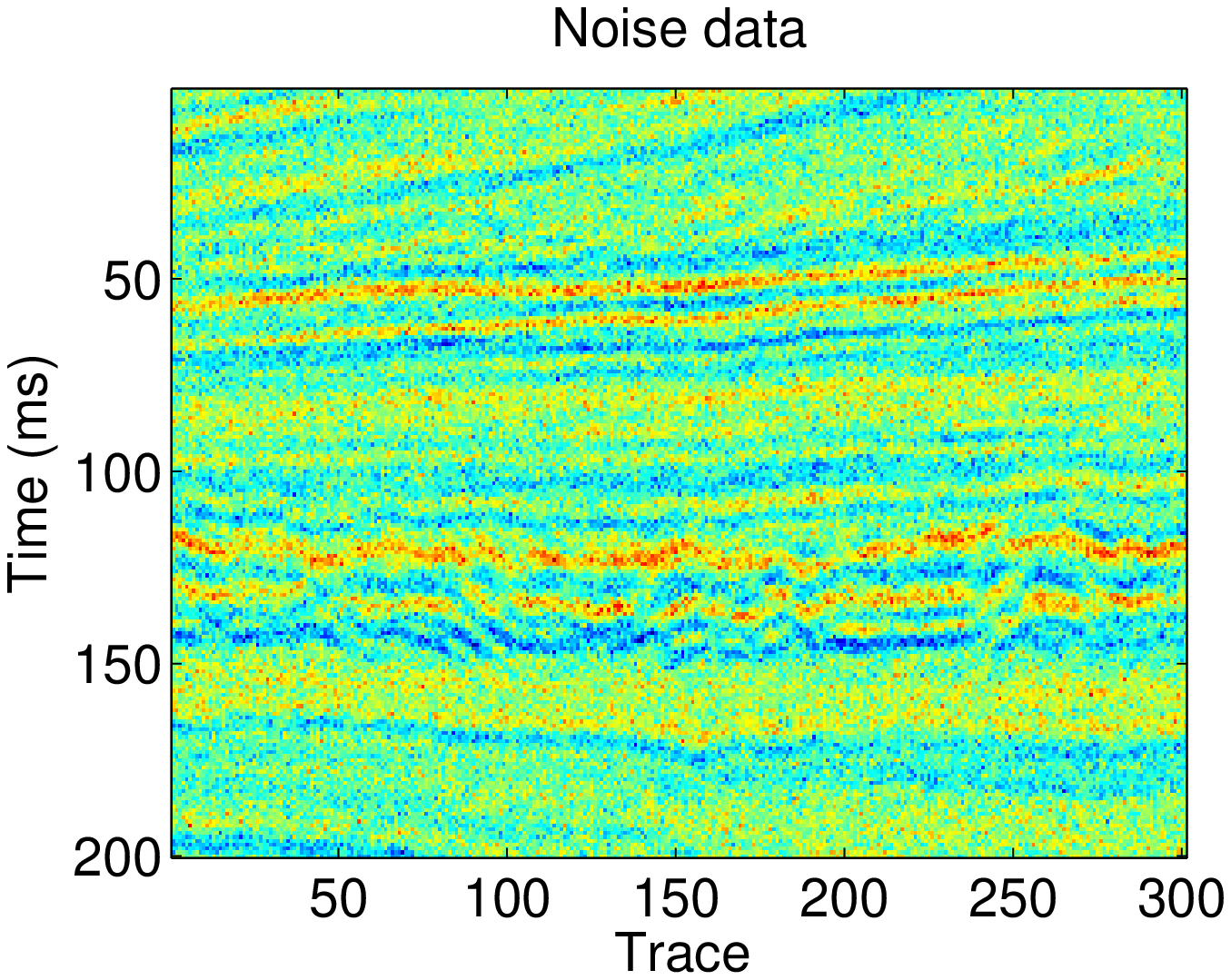}}
  \centerline{}
\end{minipage}
\caption{(a) Original data and (b) noisy data.}
\label{fig:cleandata}
\end{figure}
\subsection{Filed data example}

To verify the practicality of this method, it was applied to F3 land seismic data. The original data is shown in Figure 3(a). The simulated noisy data with Gaussian white noise is shown in Figure 3(b). The denoised results from the three different approaches are shown in Figure 4. The comparison of SNR is shown in the second row of Table 1. From the comparison of denoised results, we conclude that, while all of the three methods can obtain acceptable results, the proposed  2DSC method produces the cleanest image (Figure 4(a)) with the least amount of noisy energy left. The SNR comparison also confirms the previous observation as 2DSC-based method can obtain the highest SNR.
\section*{CONCLUSION}
In this paper, we applied a dictionary learning algorithm to seismic denoising which adopts 2DSC scheme. Alternating minimization algorithm which consists of a sparse coding learning step and a dictionary learning step  was used to solve the problem. We evaluated its denoising performance on synthetic seismic data and land data survey. The results show that this method is a good choice for suppressing noise as it can reduce computational costs and enhance the signal-to-noise ratio.
\begin{figure}[H]
\centering
\begin{minipage}{0.5\linewidth}
  \centerline{\includegraphics[width=7.5cm]{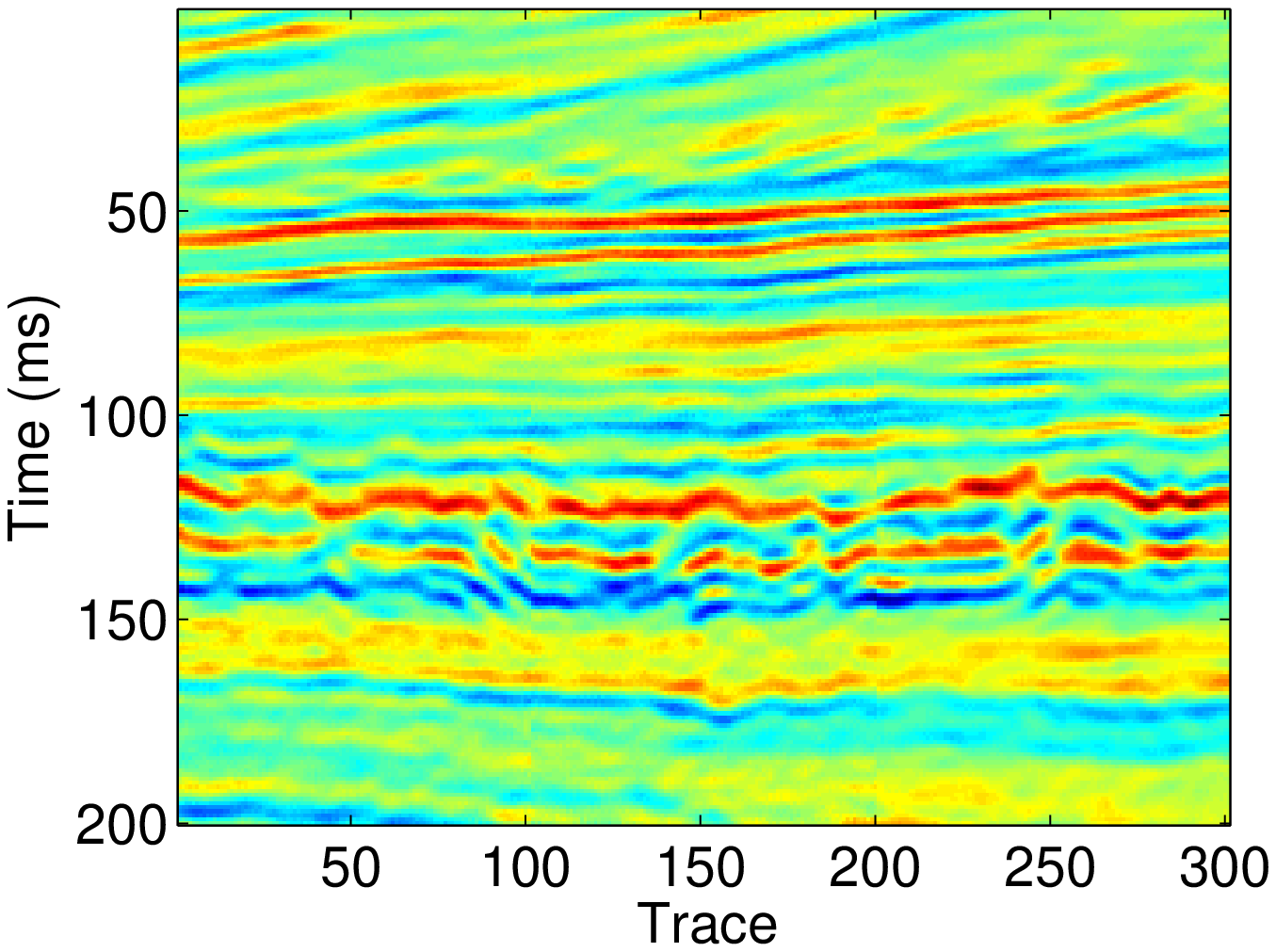}}
  \centerline{(a) 2DSC}
\end{minipage}
\begin{minipage}{0.5\linewidth}
  \centerline{\includegraphics[width=7.5cm]{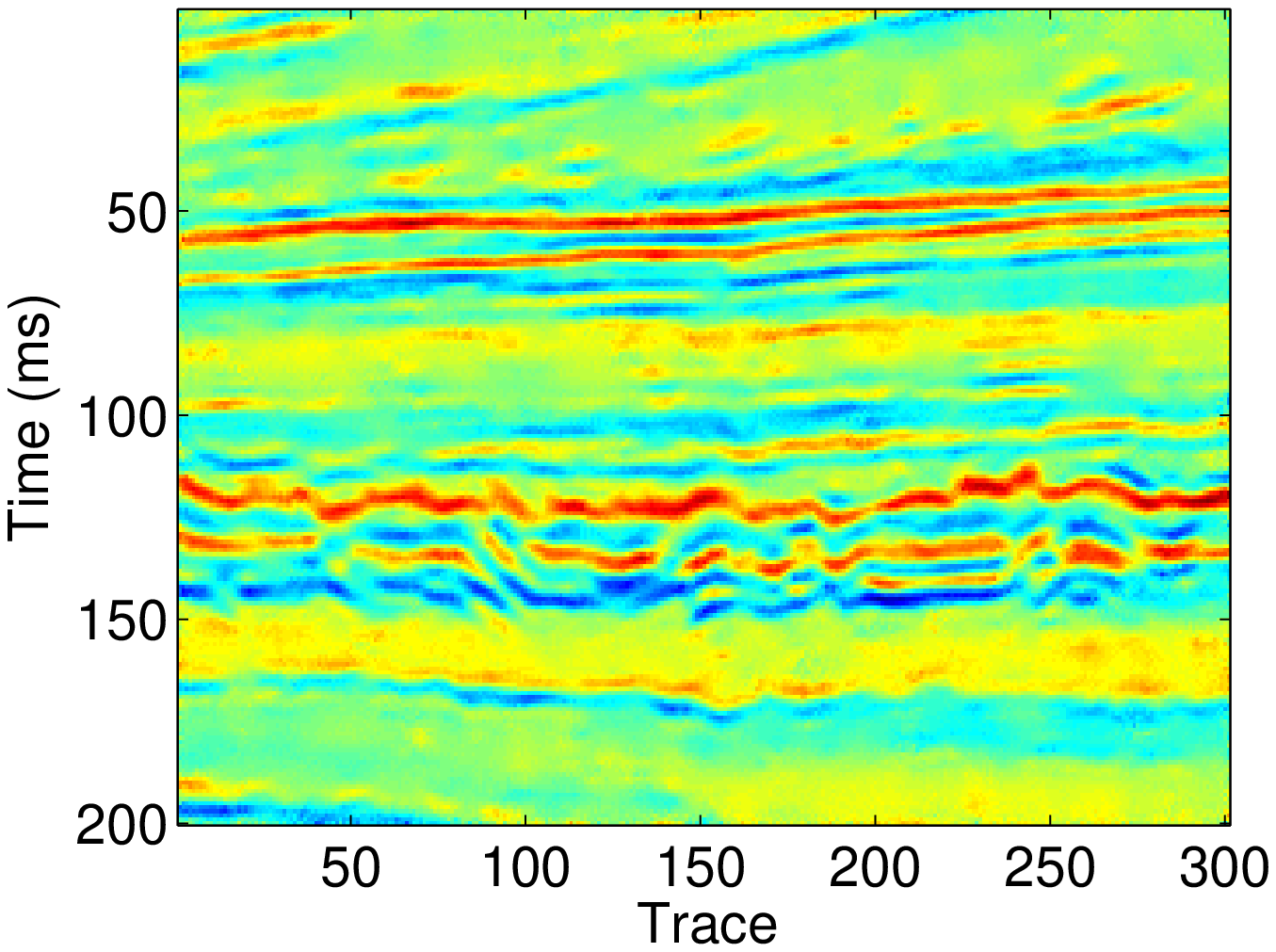}}
  \centerline{(b) K-SVD}
\end{minipage}
\begin{minipage}{0.5\linewidth}
  \centerline{\includegraphics[width=7.5cm]{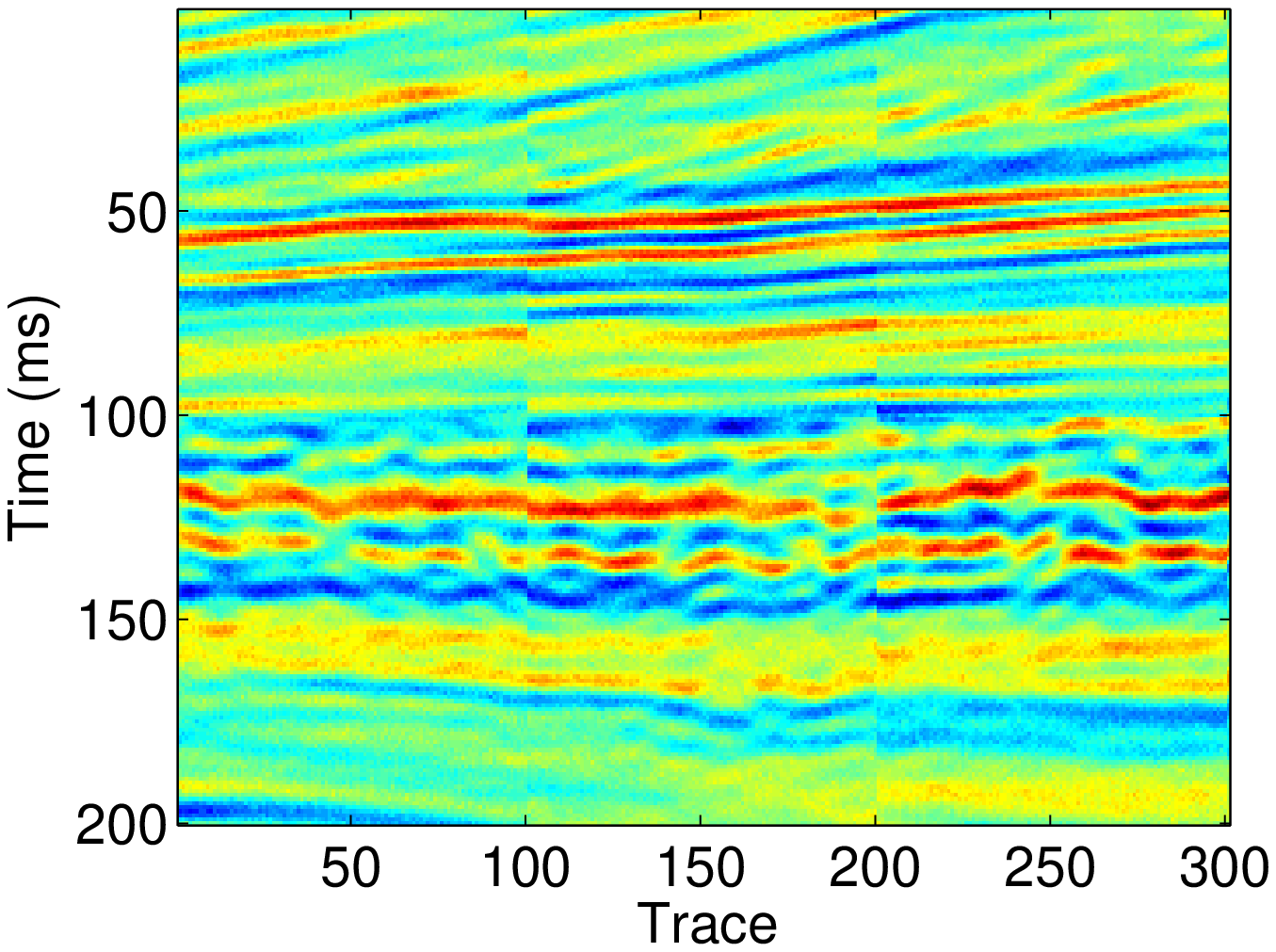}}
  \centerline{(c) MSSA}
\end{minipage}
\caption{Denoised sections using different approaches: (a) 2DSC, (b) K-SVD, and (c)  MSSA.}
\label{fig:fild}
\end{figure}
\section*{ACKNOWLEDGMENTS}
We would like to acknowledge financial support from the National Natural Science Foundation of China (Grant No.U1562218)
\onecolumn
\bibliographystyle{seg}  

\end{document}